\documentclass{bmvc2k}

\title{Guided Upsampling Network for Real-Time Semantic Segmentation}

\addauthor{Davide Mazzini}{davide.mazzini@unimib.it}{1}

\addinstitution{
 Department of Informatics, Systems and Communication\\
 University of Milano-Bicocca\\
 viale Sarca 336 Milano, Italy
}

\runninghead{Mazzini}{GUN for Real-time Semantic Segmentation}

\def\eg{\emph{e.g}\bmvaOneDot}

\def\etal{\emph{et al}\bmvaOneDot}
\def\ie{\emph{i.e}\bmvaOneDot}

\usepackage{amssymb}
\usepackage{capt-of}
\usepackage{hyperref}

\newcommand\floor[1]{\lfloor#1\rfloor}

\begin{document}

\maketitle

\begin{abstract}
Semantic segmentation architectures are mainly built upon an encoder-decoder structure. These models perform subsequent downsampling operations in the encoder. Since operations on high-resolution activation maps are computationally expensive, usually the decoder produces output segmentation maps by upsampling with parameters-free operators like bilinear or nearest-neighbor. We propose a Neural Network named Guided Upsampling Network which consists of a multiresolution architecture that jointly exploits high-resolution and large context information. Then we introduce a new module named Guided Upsampling Module (GUM) that enriches upsampling operators by introducing a learnable transformation for semantic maps. It can be plugged into any existing encoder-decoder architecture with little modifications and low additional computation cost.  We show with quantitative and qualitative experiments how our network benefits from the use of GUM module. A comprehensive set of experiments on the publicly available Cityscapes  dataset demonstrates that Guided Upsampling Network can efficiently process high-resolution images in real-time while attaining state-of-the art performances.
\end{abstract}

\section{Introduction}
Most of the current state-of-the-art architectures for image segmentation rely on an encoder-decoder structure to obtain high-resolution predictions and, at the same time, to exploit large context information. One way to increase network receptive fields is to perform downsampling operations like pooling or convolutions with large stride. Reduction of spatial resolution is twice beneficial because it also lightens the computational burden. Even state-of-the-art architectures that make use of dilated convolutions \cite{drn, psp, deeplabv3}, employ some downsampling operators in order to maintain the computation feasible. Semantic maps are usually predicted at $1/8$ or $1/16$ of the target resolution and then they are upsampled using nearest neighbor or bilinear interpolation.
\subsection{Our focus and contribution}
We focus on Semantic Segmentation of street scenes for automotive applications where a model needs to be run continuously on vehicles to take fast decisions in response to environmental events. For this reason, our design choices are the result of a trade-off between processing speed and accuracy. Our work focuses on a fast architecture with a lightweight decoder that makes use of a more effective upsampling operator. Our contributions are the following:
\begin{itemize}
  \setlength{\itemsep}{-0.1cm}
  \item We developed a novel multi-resolution network architecture named Guided Upsampling Network, presented in Section \ref{sec:network} that is able to achieve high-quality predictions without sacrificing speed. Our system can process a 512x1024 resolution image on a single GPU at 33 FPS while attaining 70.4\% IoU on the cityscapes test dataset.
  \item We designed our network in an incremental way outlining pros and cons of every choice and we included all crucial implementation details in Section \ref{sec:recipes} to make our experiments easily repeatable.
  \item We designed a novel module named GUM (Guided Upsampling Module, introduced in Section \ref{sec:gum}) to efficiently exploit high-resolution clues during upsampling.
\end{itemize}
\begin{figure}
\includegraphics[width=\columnwidth]{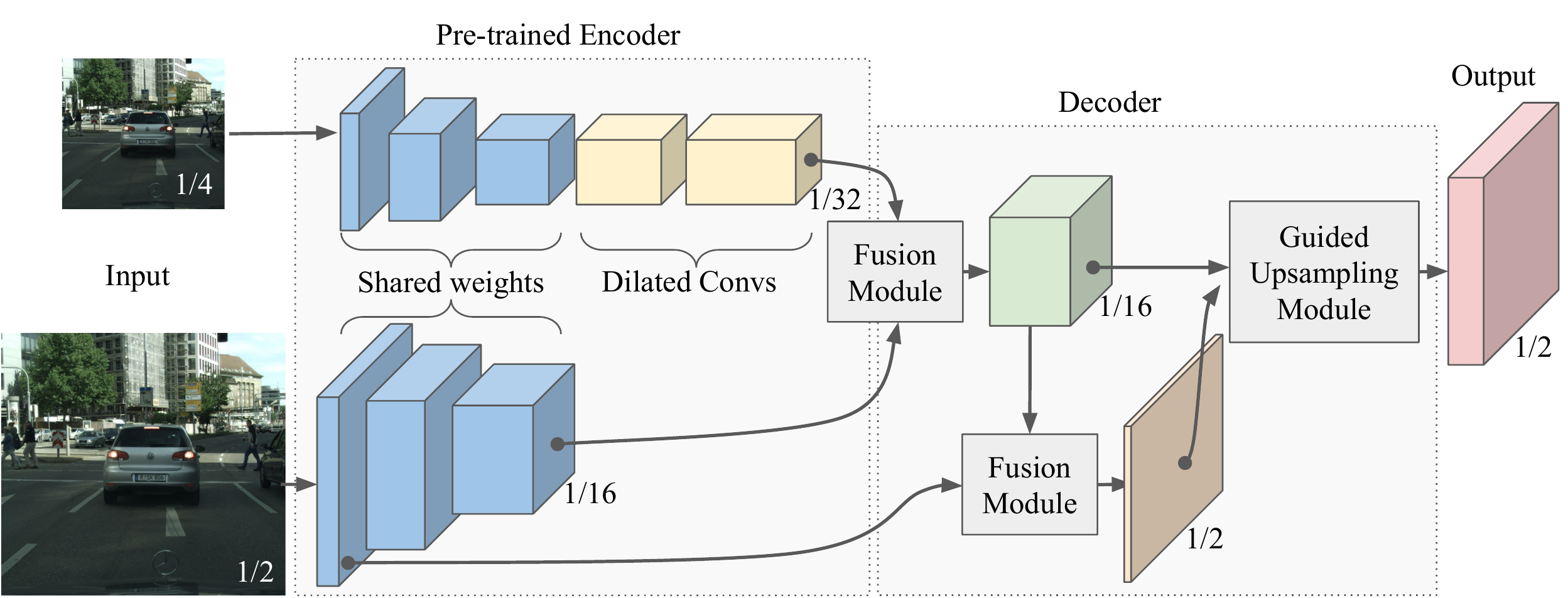}
\caption{Guided Upsampling Network (GUN) architecture. Two branches with partially shared weights extract fine and coarse features. Signals are merged by a \emph{fusion module}. Output is produced by Guided Upsampling Module (GUM) which exploits high-resolution features from earliest layers and complex features from deepest ones.}
\label{fig:architecture}
\end{figure}
\subsection{Related works}
\cite{girshick2014rich, pinheiro2014recurrent, hariharan2014simultaneous} represent the pioneer works that employed CNNs for semantic segmentation. FCN \cite{fcn} laid the foundations for modern architectures where CNNs are employed in a fully-convolutional way. Authors used a pre-trained encoder together with a simple decoder module that takes advantage of skip-connections from lower layers to exploit high-resolution feature maps. They obtained a significant improvement both in terms of accuracy and efficiency. DeepLab \cite{deeplab} made use of Dilated Convolutions \cite{dilatedconvs} to increase the receptive field of inner layers without increasing the overall number of parameters. After the introduction of Residual Networks (Resnets) \cite{resnet} most methods employed a very deep Resnet as encoder \eg DeepLabv2 \cite{deeplabv2} Resnet38 \cite{resnet38} FRRN \cite{frrn}, pushing forward the performance boundary on semantic segmentation task. PSPNet \cite{psp} and DeepLabv3 \cite{deeplabv3} introduced context layers in order to expand the theoretical receptive field of inner layers. All these methods attain high accuracy on different benchmarks but at high computational costs.

\noindent \textbf{Efficiency-oriented architectures.}
ENet authors \cite{enet} headed towards a high-speed architecture, dramatically raising model efficiency, but sacrificing accuracy. SegNet \cite{segnet} introduced an efficient way to exploit high-resolution information by saving max-pooling indices from the encoder and using them during upsampling. ICNet \cite{icnet} design is based on a three branches architecture exploiting deep supervision for training. ERFNet \cite{erfnet} implements an efficient Residual Factorized Convolution layer in order to achieve high a accuracy while being particularly efficient.

\section{Dataset and evaluation metrics}
All the experiments presented in this work have been performed on Cityscapes \cite{cityscapes}. It is a dataset of urban scenes images with semantic pixelwise annotations. It consists of 5000 finely annotated high-resolution images (2048x1024) of which 2975, 500, and 1525 belong to train, validation and test sets respectively. Annotations include 30 different object classes but only 19 are used to train and evaluate models. Adopted evaluation metrics are \emph{mean of class-wise Intersection over Union (mIoU)} and \emph{Frame Per Second (FPS)}, defined as the inverse of time needed for our network to perform a single forward pass. FPS reported in the following sections are estimated on a single Titan Xp GPU.
\section{Network design} \label{sec:network}
In this section we describe in details our network architecture. Most works in literature expose the final model followed by an ablation study. This is motivated by an implicit inductive prior towards simpler models, \ie simpler is better. Even though we agree with this line of thought we designed our experiments following a different path: by incremental steps. We started from a baseline model and incrementally added single features analyzing benefits and disadvantages. Our network architecture, based on a fully-convolutional encoder-decoder, is presented in details in the following subsections.

\noindent \textbf{Input downsampling}
A naive way to speed up inference process in real-time applications is to subsample the the input image. This comes at a price. Loss of fine details hurts performance because borders between classes and fine texture information are lost. We investigated a trade-off between system speed and accuracy. We used a DRN-D-22 model \cite{drn} pre-trained on Imagenet as encoder and a simple bilinear upsampling as decoder. First column of Table \ref{tab:multiresolution} shows the mIoU of the baseline model without any subsampling. In the second column the same model is trained and evaluated with input images subsampled by factor 4. Model speed increases from 6.7 FPS to 50.6 which is far beyond real-time but, as expected, there is a big (8\%) performance drop.
\begin{table}
\begin{center}
\resizebox{0.8\textwidth}{!}{%
\begin{tabular}{|l|c|c|c|c|c|}
\hline
Encoder & baseline & enc4 & enc24 & \textbf{enc24shared} & enc124shared\\
\hline\hline
Multiresolution & & & \checkmark & \checkmark & \checkmark\\
Shared Parameters & & & & \checkmark & \checkmark\\
\hline
Subsampling factor & 1 & 4 & 2 + 4 & \textbf{2 + 4} & 1 + 2 + 4\\
mIoU (\%) & 65.5 & 57.5 & 61.5 & \textbf{63.0} & 64.2\\
FPS & 6.7 & 50.6 & 38.7 & \textbf{38.7} & 24.9\\
\hline
\end{tabular}
}
\end{center}
\caption{Performance on Cityscapes validation set and speed (FPS) of four encoder architectures. \emph{baseline} is a full-resolution network. \emph{enc4} is trained and evaluated with downsampled input. \emph{enc24} and \emph{enc124} means 2 and 3 branches with subsampling factors 2,4 and 1,2,4 respectively. \emph{shared} means that weights are partially shared between branches. In bold the configuration adopted in the final model.}
\label{tab:multiresolution}
\end{table}

\noindent \textbf{Multiresolution encoder} \label{sec:encoder}
As a second experiment we designed a multi-resolution architecture as a good compromise to speed up the system without sacrificing its discriminative power. Our encoder consists of two branches: a low-resolution branch which is composed of all the layers of a Dilated Residual Network 22 type D (DRN-D-22) \cite{drn} with the exception of the last two. A medium-resolution branch with only the first layers of the DRN-D-22 before dilated convolutions. The main idea is to induce the first branch to extract large context features while inducing the second to extract more local features that will help to recover fine details during decoding. We experimented 3 different encoder configurations. The first named \emph{enc24} in Table \ref{tab:multiresolution} consists of two branches that process input images with sub-sampling factors 2 and 4 with the structure defined above. The second configuration named \emph{enc24shared} is similar to the first. The only difference is weight sharing between the two branches. Results in Table \ref{tab:multiresolution} show that the network with shared branches achieve better performance. We argue that, by reducing the number of network parameters, weight sharing between branches, induces an implicit form of regularization. For this reason we used this configuration as base encoder for the next experiments. In the third configuration named \emph{enc124shared} in Table \ref{tab:multiresolution} we added a further branch to elaborate full-resolution image. This indeed brought some performance improvements but we decided to discard this configuration because operations at full resolution are computationally too heavy and the whole system would slow down below the real-time threshold (30FPS). To train and evaluate the different encoder designs in Table \ref{tab:multiresolution} we fixed the decoder architecture to a configuration which is referred in Subsection \ref{sec:fusionmodule} as \emph{baseline}. Figure \ref{fig:architecture} depicts the second encoder design \emph{enc24shared}. Others have been omitted for space reasons but can be intuitively deduced.

\noindent \textbf{Fusion module.}
\label{sec:fusionmodule}
\begin{figure}
  \centering
\begin{tabular}{cc}
\bmvaHangBox{\includegraphics[height=2cm]{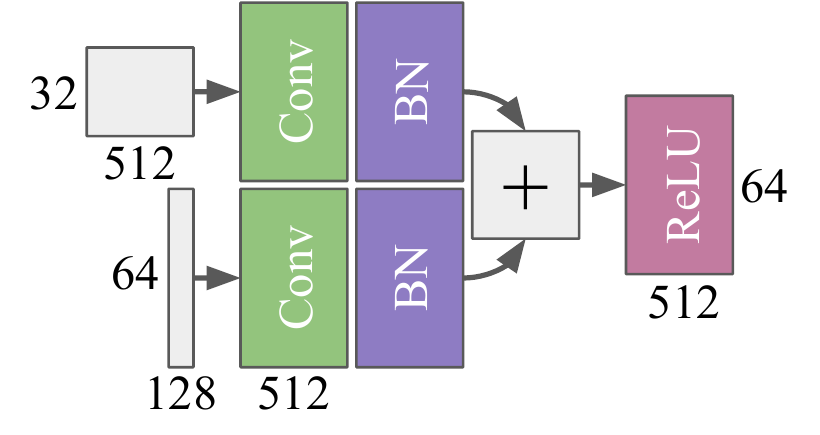}}&
\bmvaHangBox{\includegraphics[height=2cm]{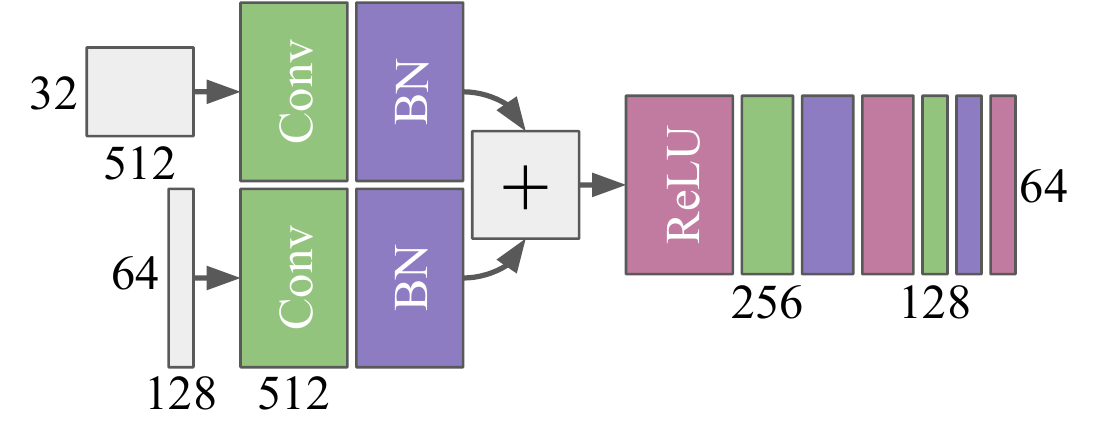}}\\
(a) base sum & (b) postproc sum
\end{tabular}
\caption{Two \emph{fusion module} configurations. Both exploit addiction as merge strategy. \emph{postproc sum} performs a dimensionality reduction.}
\label{fig:fusionmodule}
\end{figure}
\begin{table}
\begin{center}
\resizebox{0.8\textwidth}{!}{%
\begin{tabular}{|l|c|c|c|c|}
\hline
Fusion Module & base sum & base concat & \textbf{postproc sum} & postproc concat\\
\hline\hline
Sum & \checkmark & & \checkmark & \\
Concat & & \checkmark & & \checkmark \\
Postprocessing Step & & & \checkmark & \checkmark \\
\hline
mIoU (\%) & 63.0 & 63.5 & \noindent \textbf{65.8} & 64.2 \\
FPS &  38.7 &  37.8 & \textbf{37.3} & 36.4\\
\hline
\end{tabular}
}
\end{center}
\caption{mIoU on Cityscapes val set and FPS for different \emph{fusion modules}. Differences are: signal summation or concatenation and presence of a post-processing step. In bold the configuration adopted in the final model.}
\label{tab:fusionmodule}
\end{table}
It is the first part of our decoder. It joins information flows coming from the two encoder branches extracted at multiple resolutions. Input from low-resolution branch is up-sampled to match the spatial size of signal coming from the medium-resolution branch. Input coming from medium-resolution branch is expanded from 128 to 512 channels to match the number of features of the first branch. Then multi-resolution signals are merged and further processed. In Table \ref{tab:fusionmodule} are reported experimental results of four different designs. We experimented channel concatenation and addiction as merge strategies for signals coming from the two branches, named \emph{concat} and \emph{sum} respectively. We further investigated if the network benefits from feeding the final classification layer directly with the signal after the merge operation (\emph{base} in Table \ref{tab:fusionmodule}), or if a dimensionality reduction brings improvements (\emph{postproc}). From experimental results shown in Table \ref{tab:fusionmodule} both mIoU and speed take advantage of the post-processing step. The model is empowered by adding more convolutions and non-linearities and the final upsampling operations are applied to a smaller feature space. Figure \ref{fig:fusionmodule} depicts two different configurations: \emph{base sum} and \emph{postproc sum}, both with addiction merge strategy, without and with the post-processing step. Fusion modules with \emph{concat} as merge strategy have a similar structure.

\subsection{Training recipes} \label{sec:recipes}
In this section we expose our \emph{training recipes}: some considerations about hyper-parameters and their values used to train our models plus a small paragraph on synthetic data augmentation.
For all experiments in this paper we trained the network with SGD plus momentum. Following \cite{drn} we set learning rate to 0.001 and trained every model for at least 250 epochs. We adopted a \emph{step} learning rate policy. The initial value is decreased every 100 epochs by a order of magnitude. We also tried different base learning rates and \emph{poly} learning rate policy from \cite{deeplab} but we obtained better results with our baseline configuration. We found out that batch size is a very sensitive parameter affecting the final accuracy. After experimenting with different values we set it to 8. In contrast to what pointed out in \cite{atrousconvs}, increasing the batch size, in our case, hurts performance. Batch size affect performance because of intra-batch dependencies introduced by Batch Normalization layers. We argue that, in our case, the higher stochasticity introduced by intra-batch dependencies acts as regularizer, thus effectively improving the final network performance.

\noindent \textbf{Synthetic data augmentation.}
Considering the low amount of data used to train our network \ie 2970 fine-annotated images from Cityscapes dataset, we decided to investigate some well-known data augmentation techniques. The application of these techniques is almost cost-free in terms of computational resources. They do not increase processing time during inference and they can be applied as a CPU pre-processing step during training. This is in line with the research direction of this work which goal is to push forward accuracy while maintaining a real-time inference speed.
\begin{table}
\begin{center}
\resizebox{0.8\textwidth}{!}{%
\begin{tabular}{|l|c|c|c|c|}
\hline
Transformation & baseline & color jitter & lighting jitter & \textbf{random scale}\\
\hline\hline
mIoU (\%) & 65.8 & 62.6 & 64.2 & \textbf{67.5} \\
\hline
\end{tabular}
}
\end{center}
\caption{mIoU on Cityscapes validation set with different data augmentation techniques used during training. In bold the configuration adopted in the final model.}
\label{tab:dataaugmentation}
\end{table}
Since our system is supposed to work with outdoor scenes and thus dealing with a high variability of lighting conditions we experimented the use of some light transformations. Color Jitter consists in modifying image brightness, saturation and contrast in random-order. Lighting Jitter is a PCA based noise jittering from \cite{alexnet}, we used $\sigma = 0.1$ as standard deviation to generate random noise. We also experimented a geometric transform: rescaling image with a scale factor between 0.5 and 2, borrowing values from \cite{drn}. Table \ref{tab:dataaugmentation} shows the results of applying data augmentation techniques described in this section. Only \emph{random scale} brought some improvements, thus we decided to include it in our training pipeline for the next experiments.
\begin{figure}
  \centering
\begin{tabular}{cc}
\bmvaHangBox{\includegraphics[width=5.6cm]{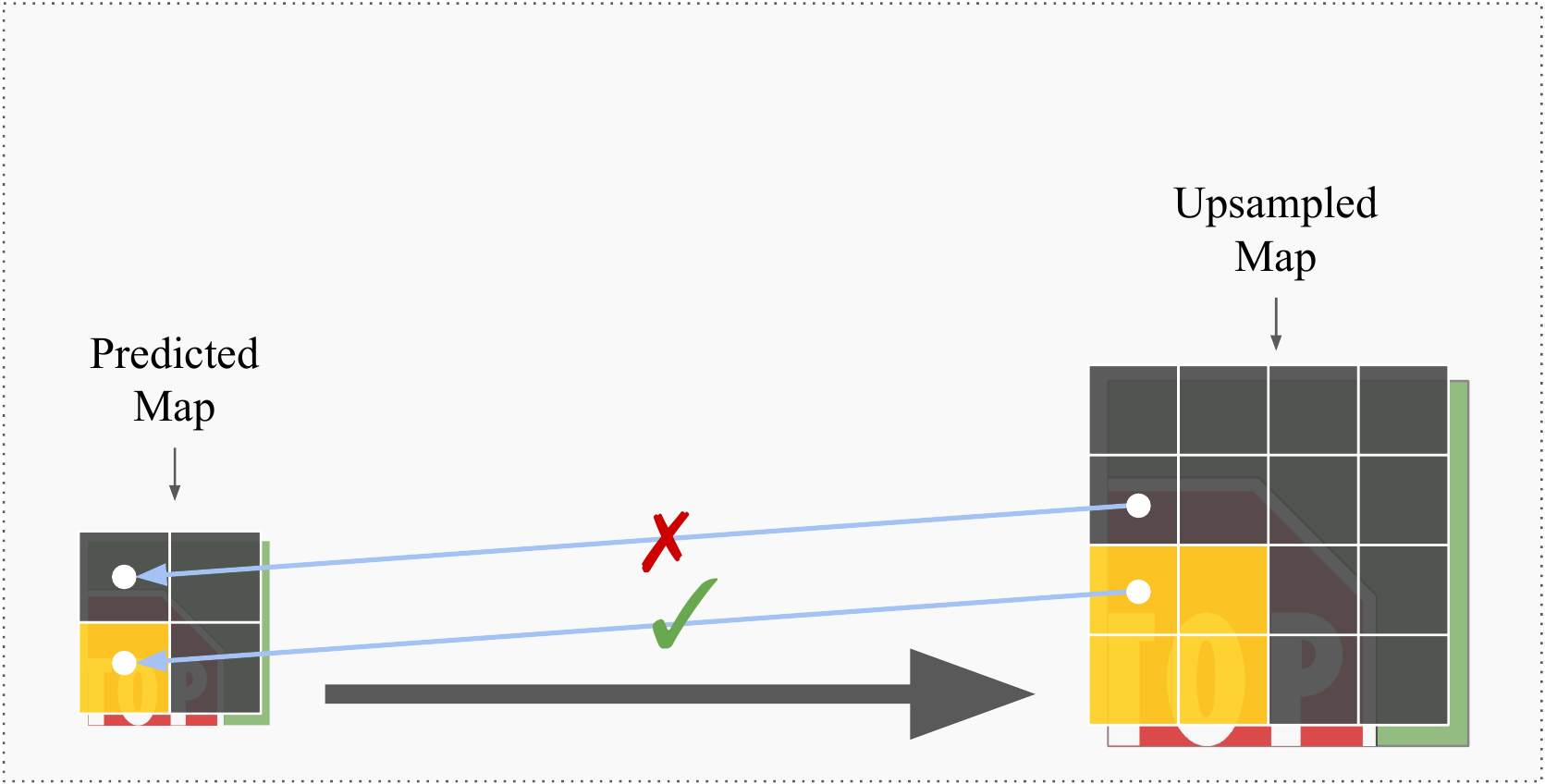}}&
\bmvaHangBox{\includegraphics[width=5.6cm]{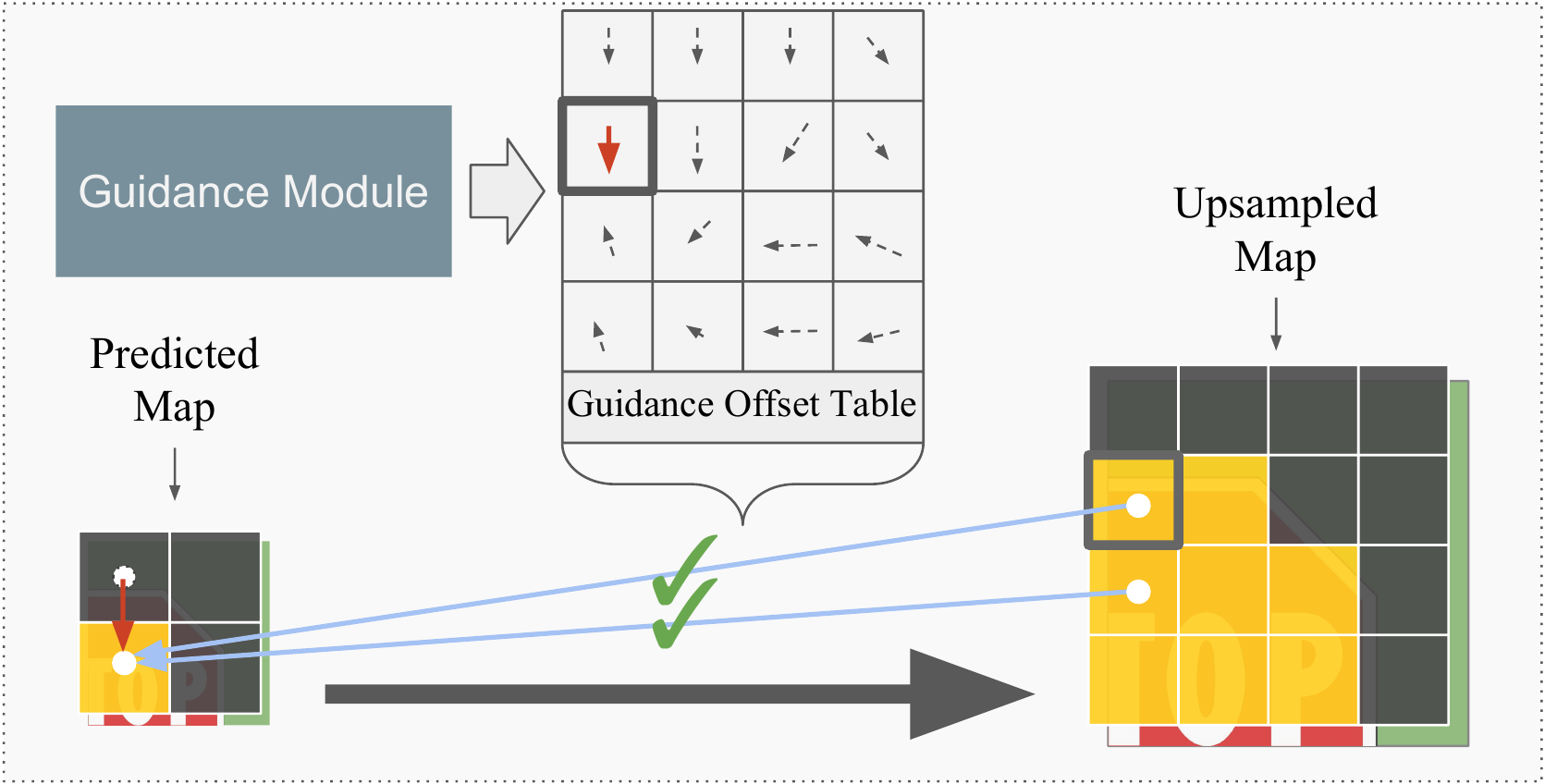}}\\
(a) Nearest Neighbor & (b) Guided Upsampling
\end{tabular}
\caption{Guided Upsampling Module (GUM). The Guidance Module produces a Guidance Offset Table which steers the upsampling process.}
\label{fig:gum}
\end{figure}
\section{Guided upsampling module} \label{sec:gum}
In this section we introduce Guided Upsampling Module (GUM). It is born from the intuition that generating a semantic map by predicting independently every single pixel is quite inefficient. It is a matter of fact that most algorithms that perform semantic segmentation do not predict full resolution maps \cite{fcn,drn, psp,icnet,deeplabv3}. They produce a low-resolution map that is up-sampled with a parameters-free operator. Usually Nearest Neighbor or Bilinear upsampling are employed. When upsampling a low-resolution map, pixels close to object boundaries are often assigned to the wrong class, see Figure \ref{fig:gum} (a). The idea behind GUM is to guide the upsampling operator through a guidance table of offsets vectors that steer sampling towards the correct semantic class. Figure \ref{fig:gum} (b) depicts the Guided Upsampling Module. A Guidance Module predicts a high-resolution Guidance Offset Table. Then GUM performs a Nearest Neighbor upsampling by exploiting the Offset Table as steering guide. Each bidimensional coordinates vector of the regular sampling grid is summed with its corresponding bidimensional vector from the Guidance Offset Table. In Figure \ref{fig:gum} the GUM module is presented in conjunction with Nearest Neighbor for simplicity, however, with simple modifications, GUM can be employed along with Bilinear operator.

Nearest Neighbor and Bilinear operators perform upsampling by superimposing a regular grid on the input feature map. Given $G_i$ the regular input sampling grid, the output grid is produced by a linear transformation $\mathcal{T}_{\theta}(G_i)$. For the specific case of upsampling, $\mathcal{T}_{\theta}$ is simply defined as:
\begin{equation}
\begin{pmatrix}x^s_i\\y^s_i\end{pmatrix} = \mathcal{T}_{\theta}(G_i) = \begin{bmatrix}\theta & 0 \\ 0 & \theta \end{bmatrix} \begin{pmatrix} x^t_i \\ y^t_i \end{pmatrix},\quad \text{with} \quad \theta \geq 1,
\end{equation}
where $(x^s_i, y^s_i) \in G_i$ are source coordinates, $(x^t_i, y^t_i)$ are target coordinates and $\theta$ represents the upsampling factor.
Given $V_i$ the output feature map and $U_{nm}$ the input feature map, GUM can be defined as follows:
\begin{equation} \label{eq:nearestguided}
  V_i = \sum^H_n{\sum^W_m{U_{nm}\delta(\floor{x^s_i + p_i + 0.5} - m)\delta(\floor{y^s_i + q_i + 0.5} - n)}}
\end{equation}
where $\floor{x^s_i + 0.5}$ rounds coordinates to the nearest integer location and $\delta$ is a Kronecker delta function. Equation \ref{eq:nearestguided} represents a sum over the whole sampling grid $U_{nm}$ where, through the Kronecker function, only a single specific location is selected and copied to the output. $p_i$ and $q_i$ represents the two offsets that shifts the sampling coordinates of each grid element in $x$ and $y$ dimensions respectively. They are the output of a function $\phi_i$ of $i$, the Guidance Module, defined as:
\begin{equation}
\phi_i = \begin{pmatrix}p_i\\q_i\end{pmatrix}
\end{equation}
Notice that $V_i$ and $U_{nm}$ are defined as bi-dimensional feature maps. The upsampling transformation is supposed to be consistent between channels therefore, equations presented in this section, generalize to multiple channels feature maps. In a similar way the bilinear sampling operator can be defined as:
\begin{equation} \label{eq:bilinearguided}
  V_i = \sum^H_n{\sum^W_m{U_{nm}\text{max}(0,1-|x^s_i + p_i - m|)\text{max}(0,1-|y^s_i + q_i - n|)}}
\end{equation}
The resulting operator is differentiable with respect to $U$ and $p_i$. We do not need the operator to be differentiable with respect to $x_i^s$ because $G_i$ is a fixed regular grid.
Equations above follows the notation used by Jaderberg \etal in \cite{stn}. In the following paragraph we will briefly outline the connection between Guided Upsampling Module and Spatial Transformer Networks.

\noindent \textbf{Connection with Spatial Transformer Networks (STN)} \cite{stn}
They introduce the ability for Convolutional Neural Networks to spatially warp the input signal with a learnable transformation. Authors of \cite{stn} separate an STN into three distinct modules: Localization Net, Grid Generator and Grid Sampler. Localization Net can be any function that outputs the transformation parameters conditioned on a particular input. Grid Generator takes as input the transformation parameters and warp a regular grid to match that specific transformation. Finally the Grid Sampler samples the input signal accordingly. Our Guided Upsampling Module can be interpreted as a Spatial Transformer Network where the Guidance Module plays the role of Localization Net and Grid Generator together. An STN explicitly outputs the parameters of a defined \emph{a priori} transformation and then applies them to warp the regular sampling grid. GUM directly outputs offsets on $x$ and $y$ directions to warp the regular sampling grid without explicitly model the transformation. Grid Sampler plays the exact same role both in GUM and STN. Since Grid Sampler module is already implemented in major Deep Learning Frameworks \eg PyTorch, TensorFlow, Caffe etc., integration of GUM within existing CNN architectures is quite straightforward.

\noindent \textbf{Guidance module.} \label{sec:guidancemodule}
The role of Guidance Module is to predict the Guidance Offset Table: the bidimensional grid that guides the upsampling process. The Guidance Module is a function which output is a tensor with specific dimensions: $HxWxC$ where $H$ and $W$ represents width and height of the high-resolution output semantic map and $C=2$ is the dimension containing the two offset coordinates w.r.t $x$ and $y$. We implemented the Guidance Module as a branch of our Neural Network, thus parameters are trainable end-to-end by backpropagation together with the whole network. We experimented three different designs for our Guidance Module and we named them \emph{large-rf}, \emph{high-res} and \emph{fusion}.
\begin{itemize}
\vspace*{-2mm}
\setlength{\itemsep}{0cm}
\item \emph{large-rf} it is composed of three upsampling layers interleaved by Conv-BatchNorm-Relu blocks. It takes the output of the fusion module and gradually upsample it. This design relies on deep network layers activations with large receptive fields but doesn't exploit high-resolution information. It is the most computationally demanding, due to the number of layers required.
\item \emph{high-res} it is composed by a single convolutional layer that takes as input a high-resolution activation map from the last Convolution before downsampling in the medium-resolution branch (see Section \ref{sec:encoder}). The convolutional layer is a 1x1 kernel and maps the 32-dimensional feature space to a 2-dimensional feature space. It is almost free in terms of computational costs because, with our architecture, it only requires 64 additional parameters and the same number of additional per-pixel operations.
\item \emph{fusion} it lies in the middle between \emph{large-rf} and \emph{high-res} modules. It merges information coming from high-resolution and large-receptive-field activation maps using the \emph{base sum} fusion module described in Section \ref{sec:fusionmodule}. It is a good compromise in terms of efficiency since it requires only two Conv-BatchNorm blocks and a single upsampling layer. Despite that, it is the one with most impact on performance because it exploit the required semantic information being at the same time faster than the iterative upsampling of \emph{large-rf} design.
\end{itemize}
\vspace*{-2mm}
Table \ref{tab:guidancemodules} reports mIoU on Cityscapes validation set and speed of the overall network in FPS. Best performance are achieved with \emph{fusion} Guidance Module.
\begin{table}
\begin{center}
\resizebox{0.6\textwidth}{!}{%
\begin{tabular}{|l|c|c|c|c|c|}
\hline
Guidance Module & baseline & large-rf & high-res & \textbf{fusion}\\
\hline\hline
Large receptive-fields & & \checkmark & & \checkmark\\
High resolution details & & & \checkmark & \checkmark\\
\hline
mIoU (\%) & 67.5 & 69.36 & 69.29 & \textbf{69.64}\\
FPS & 37.3 & 26.3 & 34.8 & \textbf{33.3}\\
\hline
\end{tabular}
}
\end{center}
\caption{mIoU and FPS with different Guidance Modules. \emph{large-rf} exploits signal from deeper layers. \emph{high-res} exploits signal from early layers. \emph{fusion} uses both. In bold the configuration adopted in the final model.}
\label{tab:guidancemodules}
\end{table}
\section{Boundaries analysis}
\begin{figure}
\centering
\begin{tabular}{cc}
  \bmvaHangBox{\includegraphics[width=0.45\textwidth]{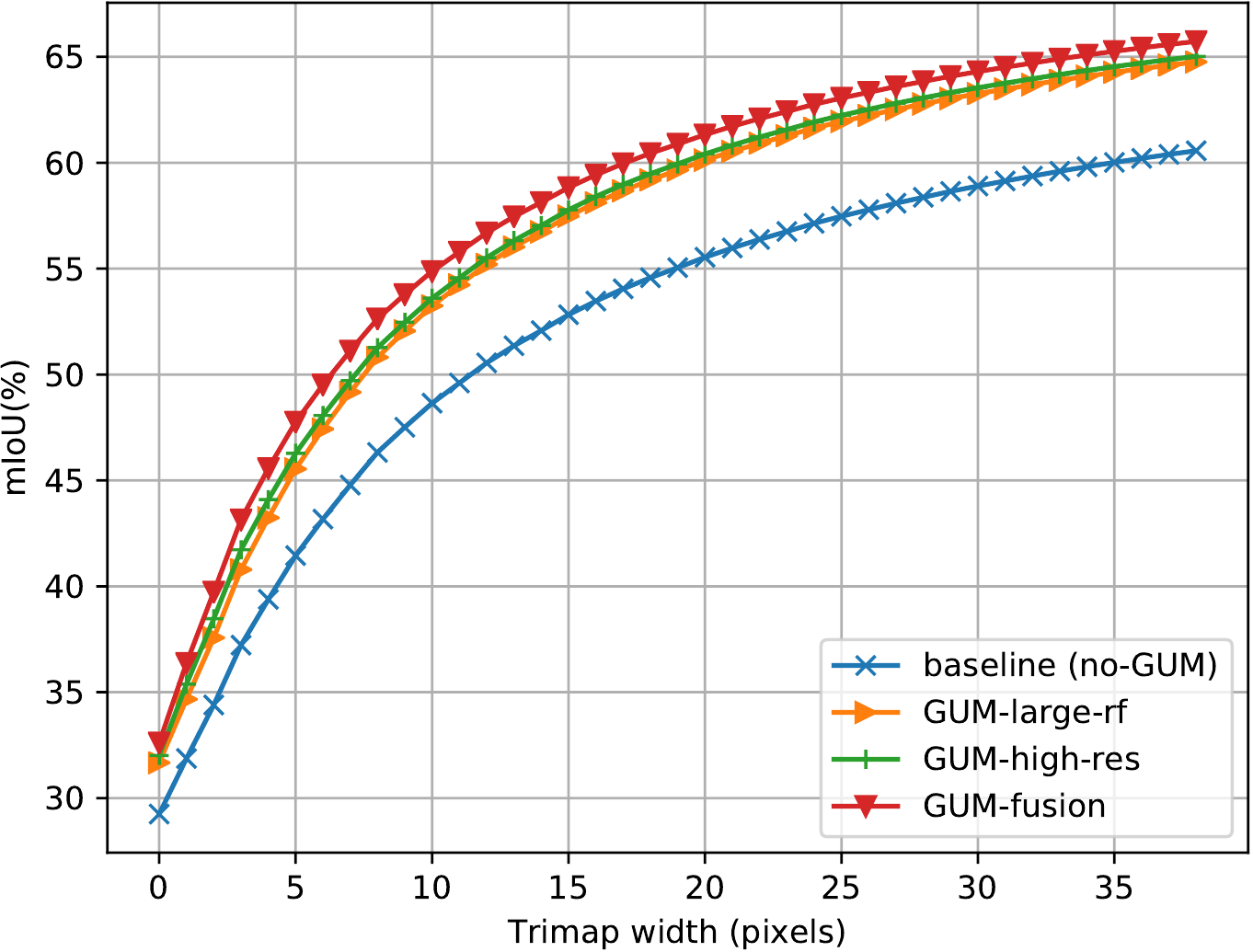}}&
  \quad
  \bmvaHangBox{\includegraphics[width=0.30\textwidth]{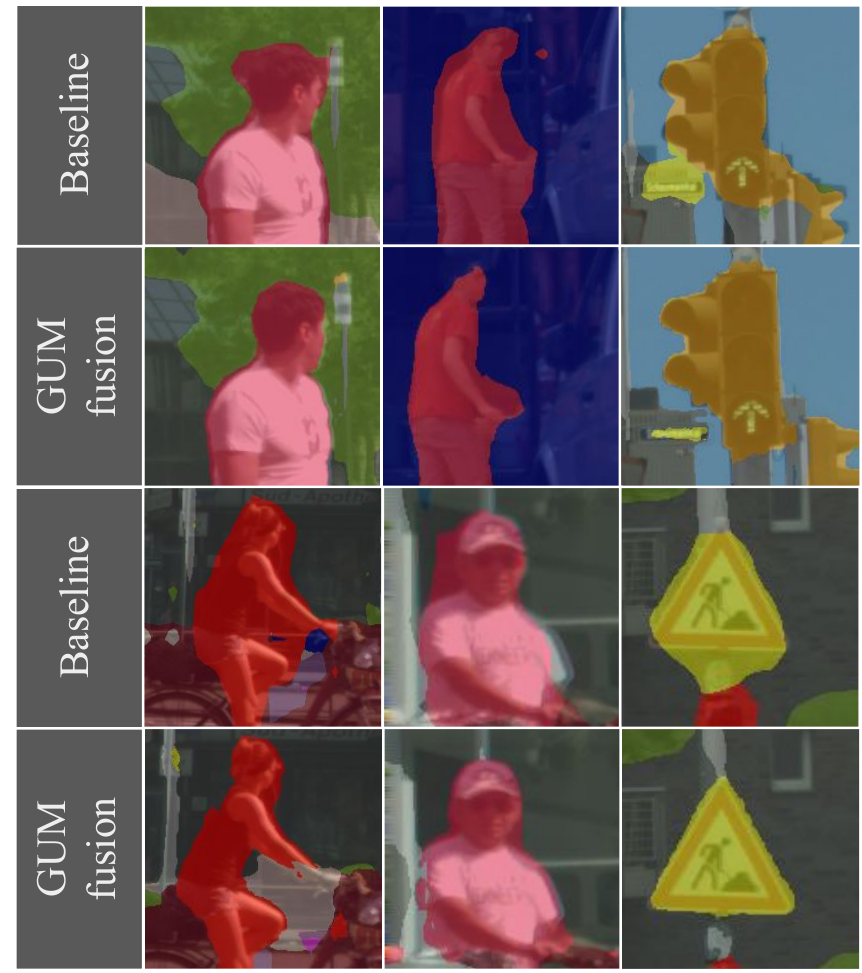}}\\
  (a) & (b)
  \end{tabular}
\caption{(a) Trimap experiment for different Guidance Modules. Major improvement is w.r.t the baseline. (b) Examples of improved boundaries with GUM.}
\label{fig:boundaries}
\end{figure}
To asses the behavior of our Guided Upsampling Network near object boundaries we performed a trimap experiment inspired by \cite{deeplab,trimap1,trimap2, deeplabv3}. The trimap experiment in \cite{deeplab, deeplabv3} was run on Pascal VOC where semantic annotations include a specific class to be ignored in train and evaluation in correspondence with object boundaries. The trimap experiment was carried out by gradually increasing annotation borders with a morphological structuring element and considering for the evaluation only pixels belonging to the expanded boundaries. To the best of our knowledge we are the first to perform the trimap experiment on Cityscapes dataset. Since there is no boundary class to expand we decided to implement the experiment in a different but equivalent way: for each object class independently we computed the distance transform on ground-truth maps. Then we performed the trimap experiment by gradually increasing the threshold on our computed distance transform map to include pixels at different distances from object boundaries. Figure \ref{fig:boundaries} shows a qualitative and quantitative comparison of the three Guidance Modules \ie \emph{large-rf}, \emph{high-res} and \emph{fusion} with respect to the \emph{baseline}, where baseline is the exact same network with bilinear upsampling instead of GUM. There is a clear advantage in using GUM versus the baseline. The type of Guidance Module does not drastically affect the results even though GUM with \emph{fusion} achieve slightly higher mIoU levels.

\section{Comparison with the state-of-the-art}
In Table \ref{tab:comparison} we reported performance of Guided Upsampling Network along with state-of-the-art methods on Cityscapes test set. Segmentation quality has been evaluated by Cityscapes evaluation server and it is reported in the official leaderboard\footnote {https://www.cityscapes-dataset.com/benchmarks/}. FPS in Table \ref{tab:comparison} have been estimated on a single Titan Xp GPU. For fairness we only included algorithms that declare their running time on Cityscapes leaderboard, even though DeepLabv3+\cite{deeplabv3} has been listed in Table \ref{tab:comparison} as a reference for accuracy-oriented methods. Usually, methods that do not care about processing time, are computationally heavy. Most of them \eg PSPNet, DeepLabv3 \cite{psp,deeplabv3} achieve very high mIoU levels, \ie DeepLabv3+ is the best published model to date, reaching 81.2\%, but they adopt very time-consuming multi-scale testing to increase accuracy. Our Guided Upsampling Network achieve 70.4\% of mIoU on Cityscapes test set without any postprocessing. To the best of our knowledge this is the highest mIoU for a published method running at >30 FPS. It performs even better than some methods like Adelaide, Dilation10 etc. that do not care about speed.
\begin{table}
  \centering
\begin{minipage}[b]{0.50\textwidth}
  \centering
\resizebox{0.98\textwidth}{!}{%
\begin{tabular}{|l|c c c|}
\hline
Name & Subsampling & mIoU(\%) & FPS\\
\hline\hline
SegNet \cite{segnet} & 4 & 57.0 & 26.4 \\
ENet \cite{enet} & 2 & 58.3 & 121.5 \\
SQ \cite{sq} & no & 59.8 & 26.4 \\
CRF-RNN \cite{crfrnn} & 2 & 62.5 & 2.2 \\
DeepLab \cite{deeplab} & 2 & 63.1 & 0.4 \\
FCN-8S \cite{fcn} & no & 65.3 & 4.9 \\
Adelaide \cite{adelaide} & no & 66.4 & 0.05 \\
Dilation10 \cite{dilatedconvs} & no & 67.1 & 0.4 \\
ICNet \cite{icnet} & no & 69.5 & 47.9 \\
ERFNet \cite{erfnet} & 2 & 69.7 & 52.6\\
\textbf{GUN (ours)} & \textbf{2} & \textbf{70.4} & \textbf{33.3} \\
DeepLabv3+\cite{deeplabv3} & no & 81.2 & n\textbackslash a \\
\hline
\end{tabular}
}
\vspace*{2mm}
\caption{Comparison with state-of-the-art methods on Cityscapes test set sorted by increasing mIoU. Our method in boldface.}
\label{tab:comparison}
\end{minipage}
\hfill
\begin{minipage}[b]{0.47\textwidth}
\centering
\includegraphics[width=0.915\textwidth]{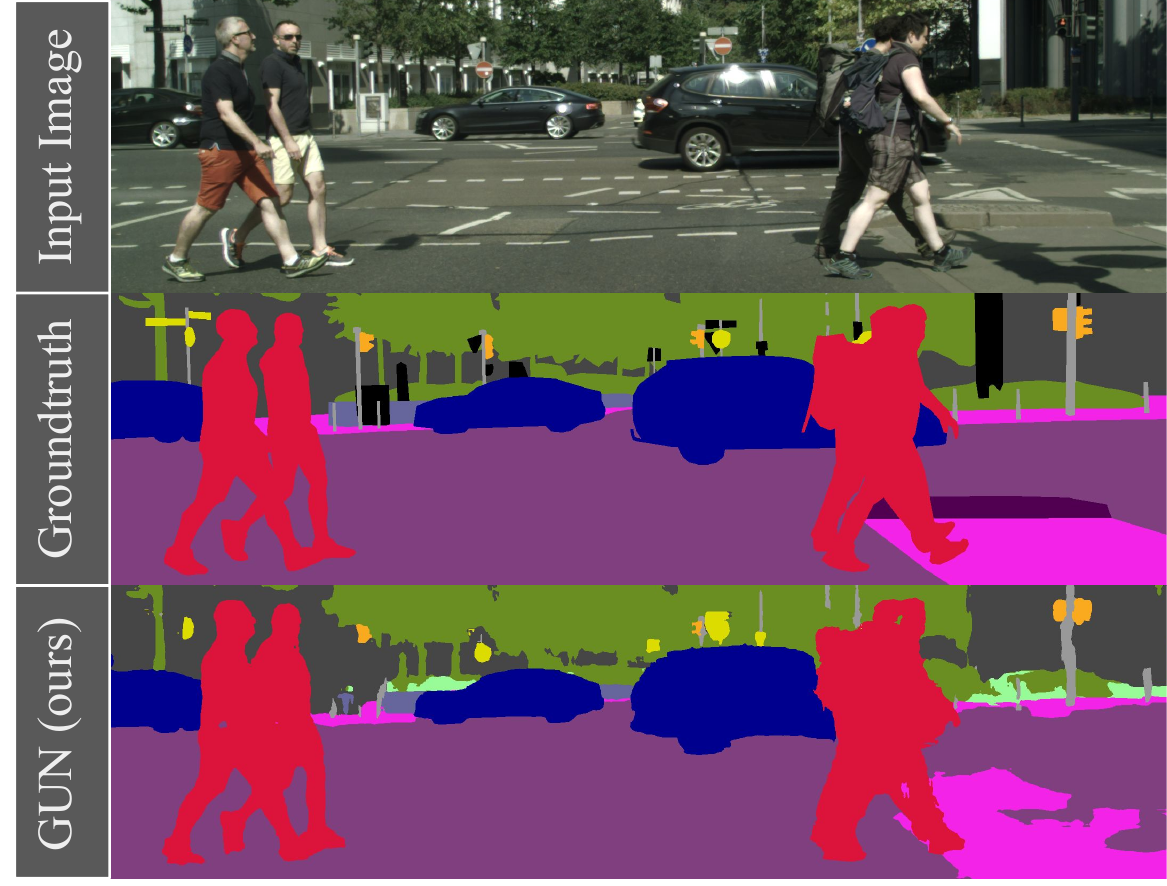}
\vspace*{-2mm}
\captionof{figure}{From top to bottom respectively input image, ground-truth and prediction obtained with our Guided Upsampling Net.}\label{fig:segmentation}
\label{fig:image}
\end{minipage}
\end{table}
\section{Conclusions}
We proposed a novel network architecture to perform real-time semantic segmentation of street scene images. It consists of a multiresolution architecture to jointly exploit high-resolution textures and large context information. We introduced a new module named Guided Upsampling Module to improve upsampling operators by learning a transformation conditioned on high-resolution details. We included GUM in our network architecture and we experimentally demonstrated performance improvements with low additional comptutational costs. We evaluated our network on the Cityscapes test dataset showing that it is able to achieve 70.4\% mIoU while running at 33.3 FPS on a single Titan Xp GPU. Further details and a demo video can be found in our project page: \url{http://www.ivl.disco.unimib.it/activities/semantic-segmentation}.
\section{Acknowledgements}
The research leading to these results has received funding from TEINVEIN: TEcnologie INnovative per i VEicoli Intelligenti, Unique Project Code: E96D17000110009 - Call "Accordi per la Ricerca e l'Innovazione", cofunded by POR FESR 2014-2020 Regional Operational Programme, European Regional Development Fund.\\
We gratefully acknowledge the support of NVIDIA Corporation with the donation of a Titan Xp GPU used for this research.

\bibliography{semseg}
\end{document}